\documentclass[letterpaper]{article} 
\usepackage{aaai2027}  
\usepackage[hyphens]{url}  
\usepackage{graphicx} 
\usepackage{natbib}  
\usepackage{caption} 
\usepackage{algorithm}
\usepackage{algorithmic}

\usepackage{newfloat}
\usepackage{listings}
\DeclareCaptionStyle{ruled}{labelfont=normalfont,labelsep=colon,strut=off} 
\floatstyle{ruled}
\newfloat{listing}{tb}{lst}{}
\floatname{listing}{Listing}

\usepackage{booktabs}
\usepackage{amssymb}
\usepackage{array}
\usepackage{colortbl}
\definecolor{caseheader}{RGB}{239,242,247}
\definecolor{casepanel}{RGB}{248,249,251}
\definecolor{caseborder}{RGB}{198,205,216}
\definecolor{caseblue}{RGB}{226,239,250}
\definecolor{casegreen}{RGB}{226,244,235}
\definecolor{casered}{RGB}{252,230,230}

\title{Evaluating Counterfactual Sensitivity to Patient Information in Medication-Safety Reasoning}
\author {
    Zhitian Hou\textsuperscript{\rm 1, \rm 2},
    Yuhang Liu\textsuperscript{\rm1, \rm 2},
    Pengkai Wang\textsuperscript{\rm 1},
    Zeyu Liu\textsuperscript{\rm 1}, 
    Guanghao Zhu\textsuperscript{\rm 1},
    Zheng Liu\textsuperscript{\rm 1, \rm 2},
    Shuo Cai\textsuperscript{\rm 1},
    Congkai Xie\textsuperscript{\rm 2},
    Zhijie Sang\textsuperscript{\rm 2},
    Kun Zeng\textsuperscript{\rm 3},
    Hongxia Yang\textsuperscript{\rm 1, \rm 2}\corresponding
}
\affiliations {
    \textsuperscript{\rm 1}The Hong Kong Polytechnic University\\
    \textsuperscript{\rm 2}InfiX.ai\\
    \textsuperscript{\rm 3}Sun Yat-sen University\\
    zhitian.hou@connect.polyu.hk, hongxia.yang@polyu.edu.hk
}

\begin{document}

\maketitle

\begin{abstract}
Applying a valid medication-safety rule when its patient-specific conditions are not met can produce an incorrect decision. Existing medical evaluations largely use isolated and fixed scenarios. A model may therefore answer correctly by recalling a drug-risk association without showing that it used patient information to decide whether the rule applies. To address this gap, we introduce MedPIC-Bench, a benchmark of source-verifiable recommendations and expert-validated questions for patient-specific medication-safety reasoning. It combines guideline-following questions with paired counterfactual questions in which a controlled change in patient information changes whether a rule applies. The benchmark contains 467 questions annotated along six clinical and reasoning dimensions. Across 28 medical-specific, general, and proprietary LLMs, every model performs worse on counterfactual questions, with mean accuracy falling from 63.6\% to 45.1\%. Models perform well when an explicit patient attribute directly signals a familiar contraindication, but struggle when patient information must narrow or withdraw a safety warning. Model rationales often acknowledge the changed patient information, yet the final answers retain the previous safety judgment. This vulnerability persists among medical-specific LLMs, whose average CF performance trails that of general LLMs. MedPIC-Bench therefore makes conditional rule application measurable and highlights the limitations of static medication-safety accuracy for assessing patient-specific reliability.
\end{abstract}


\section{Introduction}

\begin{figure}[!t]
    \centering
    \includegraphics[width=\columnwidth]{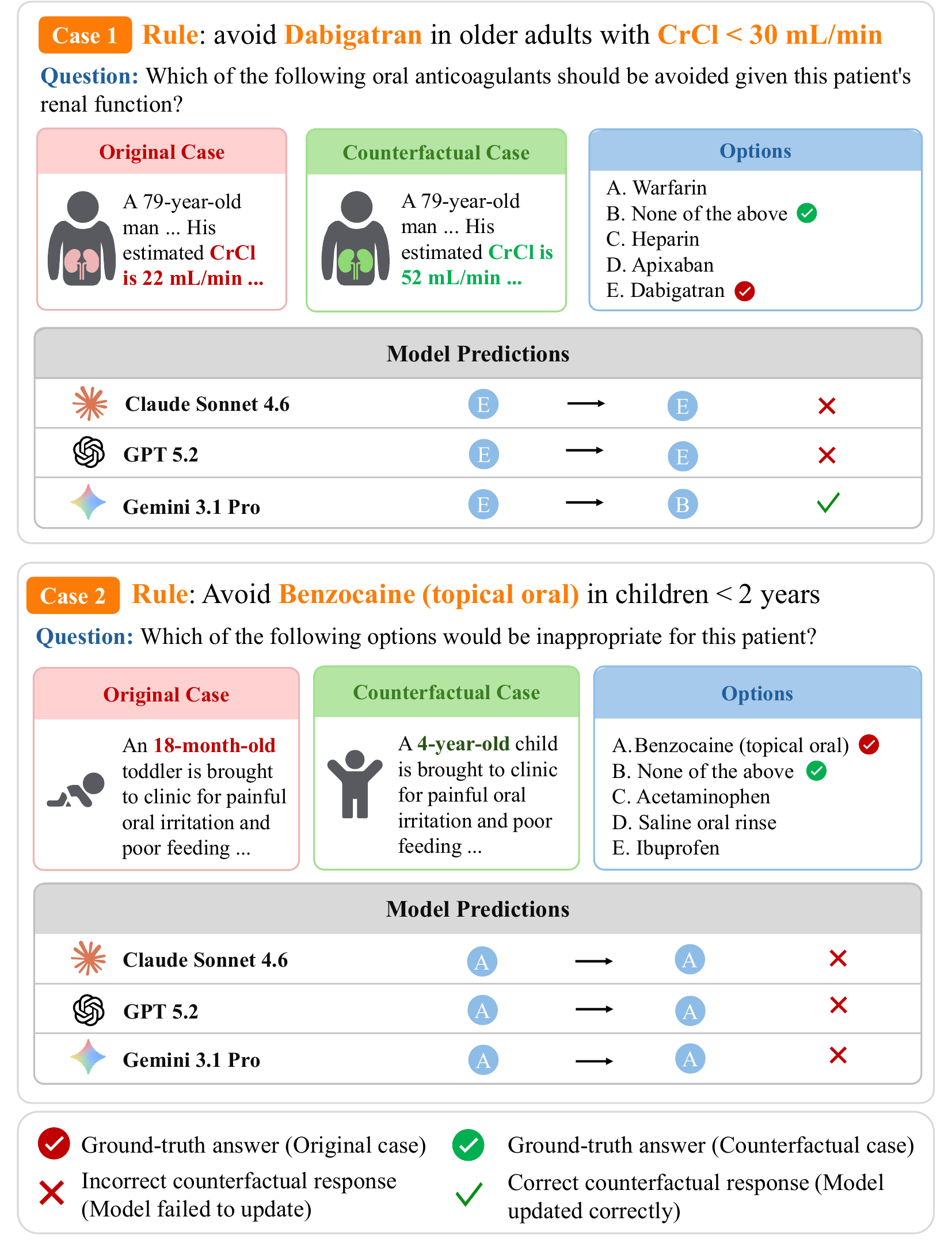}
    \caption{Two paired counterfactual examples from MedPIC-Bench. A change in patient information reverses whether the safety rule applies, yet strong models often retain the original risk judgment.}
    \label{fig:counterfactual-examples}
\end{figure}

Large language models (LLMs) are increasingly evaluated for medical decision-making. Existing benchmarks range from broad medical QA settings such as MultiMedQA \cite{singhal2023large} and Med-PaLM \cite{singhal2025toward} to open-ended clinical assessment frameworks \cite{arora2025healthbench}. However, most medical benchmarks use fixed patient scenarios. They test whether a model can identify a risky medication in a given case, but not whether it revises that judgment when relevant patient information changes. In real prescribing, one variable, such as age, syndrome subtype, or comorbidity, can change the appropriate judgment. A model can therefore succeed on a static question while failing to revise its answer when that variable changes, as illustrated in Figure~\ref{fig:counterfactual-examples}.

This distinction matters because medication-safety rules are conditional. A drug may be inappropriate within a specific age range, below a renal threshold, or in the presence of a particular clinical condition. If evaluation presents only one fixed scenario, strong performance may reflect a memorized drug-risk association rather than correct use of the condition that makes the rule applicable. Recent benchmarks have moved beyond exam-style medical QA. MedGuide \cite{li2025medguide} evaluates guideline-based clinical decisions, RxSafeBench \cite{zhao2025rxsafebench} tests medication hazards in simulated consultations, and MedEinst \cite{chen2026medeinst} probes diagnostic fixation through paired changes in clinical evidence. However, they do not test whether models revise their medication-safety judgments when patient information changes whether the underlying rule applies.

MedPIC-Bench addresses this gap through source-verifiable recommendations and expert-validated questions for patient-specific medication-safety reasoning. It combines \textbf{guideline-following (GF)} questions with paired \textbf{counterfactual (CF)} questions. GF questions test whether a model follows the applicable medication-safety recommendation in a fixed patient case. In each CF pair, a controlled change in patient information changes whether the underlying rule applies, while the remaining case details and answer options are kept fixed whenever possible. MedPIC-Bench contains 467 questions spanning 9 organ systems, 11 clinical departments, 66 drug categories, 8 patient-information types, 3 special populations, and 5 reasoning operations, enabling fine-grained analysis of where models fail.

Across 28 medical-specific, general, and proprietary LLMs, mean accuracy falls from 63.6\% on GF questions to 45.1\% on CF questions, with every model declining. Paired analysis shows that the deficit is not limited to recalling medication risks. Models frequently preserve a risk judgment after the patient information that triggered it has been removed, even when their rationales acknowledge the change. A within-family comparison indicates that medical adaptation can improve absolute accuracy, but the persistent deactivation gap shows that it does not resolve the underlying failure. These results expose a gap between recognizing a medication risk and controlling whether that risk applies to the patient at hand.

Our main contributions are as follows.
\begin{itemize}
    \item We frame counterfactual sensitivity to patient information as a distinct requirement for medication-safety reasoning. Within this framing, we identify a recurring failure pattern in which models recognize relevant patient information but fail to let it constrain or withdraw a familiar drug-risk judgment.
    \item We introduce \textbf{MedRule2Pair}, a source-grounded construction pipeline that transforms heterogeneous clinical recommendations into verified, traceable rules and controlled GF and CF question variants, supporting reproducible evaluation and expansion.
    \item We instantiate this pipeline as \textbf{MedPIC-Bench}, an expert-validated benchmark of 467 questions annotated along six dimensions. Across 28 evaluated models, every model declines from GF to CF. Medical-specific LLMs also do not outperform general LLMs on average, achieving a mean CF accuracy of 38.3\%, compared with 45.5\% for general LLMs.
\end{itemize}

\section{Related Work}

\textbf{Medical LLM and guideline-based evaluation.} Medical LLM evaluation has expanded from exam-style question answering to broader assessments of clinical response quality and decision-making. MultiMedQA~\cite{singhal2023large} and the Med-PaLM studies~\cite{singhal2025toward} evaluate medical knowledge and reasoning across multiple question-answering settings, while HealthBench~\cite{arora2025healthbench} evaluates open-ended responses to realistic health conversations. MultifacetEval~\cite{zhou2024multifaceteval} probes medical knowledge from several complementary perspectives, and MedGuide~\cite{li2025medguide} evaluates clinical decisions against guideline-derived criteria. MedGuideX~\cite{shen2026medguidex} goes further by transforming executable guideline logic into factual and counterfactual supervision for model post-training. These efforts broaden medical evaluation and guideline-based reasoning, but they do not provide a dedicated benchmark for measuring whether source-grounded medication-safety rules are applied and withdrawn appropriately across controlled patient cases.

\textbf{Medication-safety and prescribing benchmarks.} Recent benchmarks have begun to evaluate medication use more directly. Rx-LLM~\cite{zhao2025rx} introduces clinician-annotated tasks covering medication knowledge and safety-critical pharmacy operations. RxSafeBench~\cite{zhao2025rxsafebench} embeds contraindications and drug interactions in simulated consultation scenarios to test safe medication selection, while RxEval~\cite{chen2026rxeval} presents detailed patient profiles and longitudinal clinical trajectories for prescription-level medication recommendation. These efforts move beyond generic medical QA and show that models can struggle to integrate medication knowledge with patient information. Their evaluation units, however, remain individual tasks, consultations, or prescriptions. Consequently, a correct response does not by itself establish whether the relevant patient information governed the decision.

\textbf{Counterfactual evaluation in clinical reasoning.} CounterBench~\cite{chen2026counterbench} evaluates formal counterfactual reasoning over synthetic causal structures. In medicine, MediEval~\cite{qu2026medieval} jointly tests knowledge grounding and contextual consistency using factual and counterfactual statements linked to patient records. Counterfactual Patient Variations~\cite{benkirane2025diagnose} and MEDEQUALQA~\cite{ghosh2025medequalqa} use controlled demographic changes to audit bias and reasoning stability, whereas MedEinst~\cite{chen2026medeinst} changes discriminative evidence across paired cases to expose diagnostic fixation. MedPIC-Bench instead asks whether models update medication-safety judgments when patient information changes the applicability of a verified prescribing rule. By combining guideline-following evaluation with controlled counterfactual pairs, it measures both rule application in individual cases and whether decisions change in the correct direction across matched cases.

\section{MedPIC-Bench Construction}

\begin{figure*}[t]
    \centering
    \includegraphics[width=\textwidth]{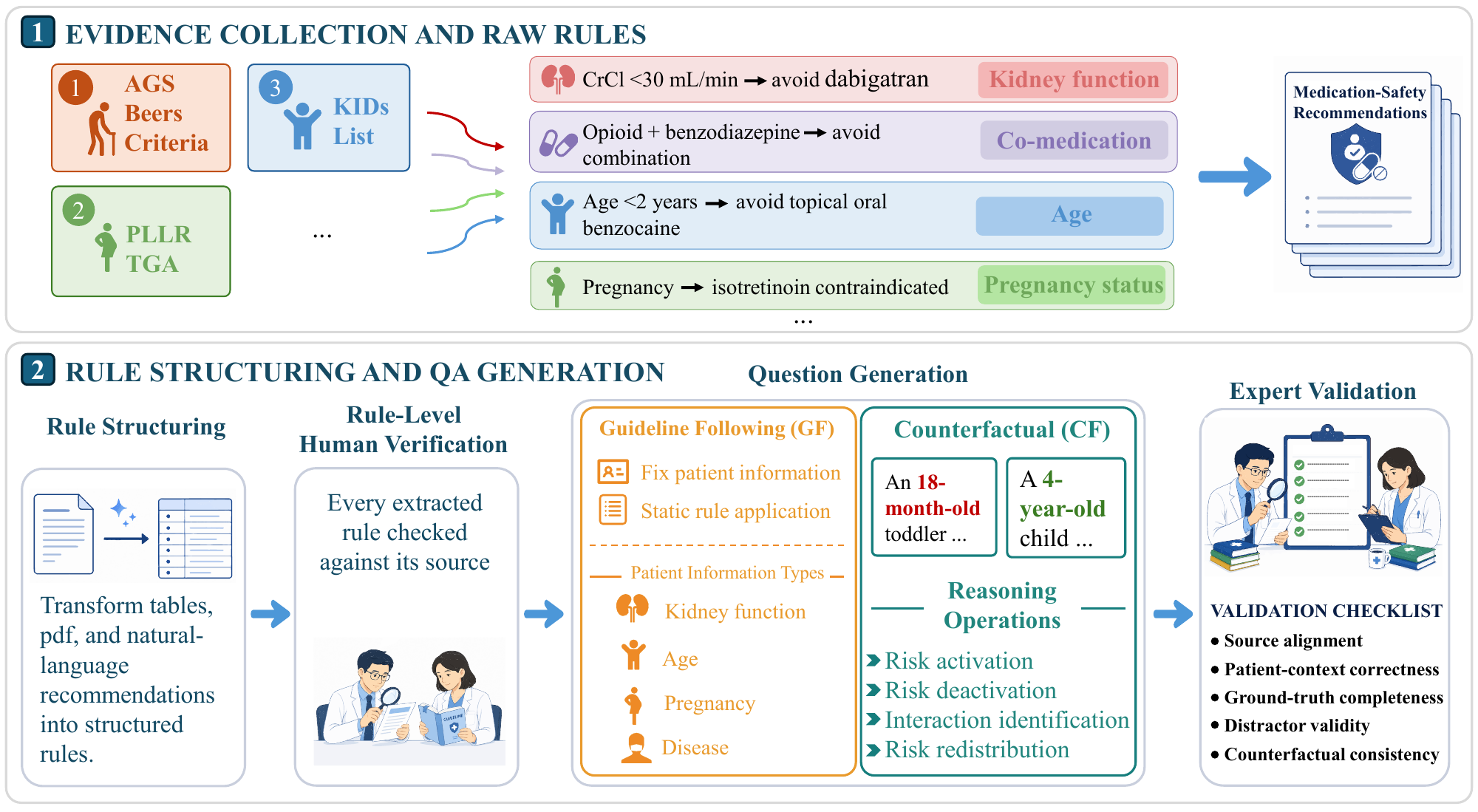}
    \caption{Overview of MedRule2Pair, our two-stage pipeline for constructing source-grounded medication-safety questions. First, recommendations collected from authoritative sources yield raw rules linked to explicit patient information. Second, these rules are structured, manually verified against their sources, and used to generate GF questions with fixed patient information and CF questions with controlled changes, followed by expert validation. The pipeline preserves source traceability while isolating the information that changes whether a rule applies.}
    \label{fig:construction}
\end{figure*}

MedPIC-Bench is built around two design requirements. First, every answer must be traceable to a verified medication-safety recommendation. Second, each counterfactual comparison must isolate a targeted change in patient information. We implement these requirements through MedRule2Pair, a two-stage construction pipeline shown in Figure~\ref{fig:construction}. Source recommendations are first converted into structured rules and manually verified, and the verified rules are then used for question generation and expert validation.

\subsection{Evidence Collection and Raw Rules}

We compile medication-safety recommendations from authoritative resources, including the American Geriatrics Society Beers Criteria \cite{20232023american}, the Pregnancy and Lactation Labeling Rule (PLLR)\footnote{\url{https://www.fda.gov/drugs/labeling-information-drug-products/pregnancy-and-lactation-labeling-drugs-final-rule}}, the Australian Therapeutic Goods Administration (TGA) Prescribing Medicines in Pregnancy Database\footnote{\url{https://www.tga.gov.au/resources/health-professional-information-and-resources/australian-categorisation-system-prescribing-medicines-pregnancy/prescribing-medicines-pregnancy-database}}, and the 2025 KIDs List \cite{mcpherson2025pediatric}. From these sources, we retain recommendations that state a verifiable medication-safety decision and make its applicability depend on explicit patient information, such as kidney function, concomitant medication use, age, or pregnancy status.

\subsection{Rule Structuring and Question Generation}

\begin{figure*}[t]
    \centering
    \includegraphics[width=0.8\textwidth]{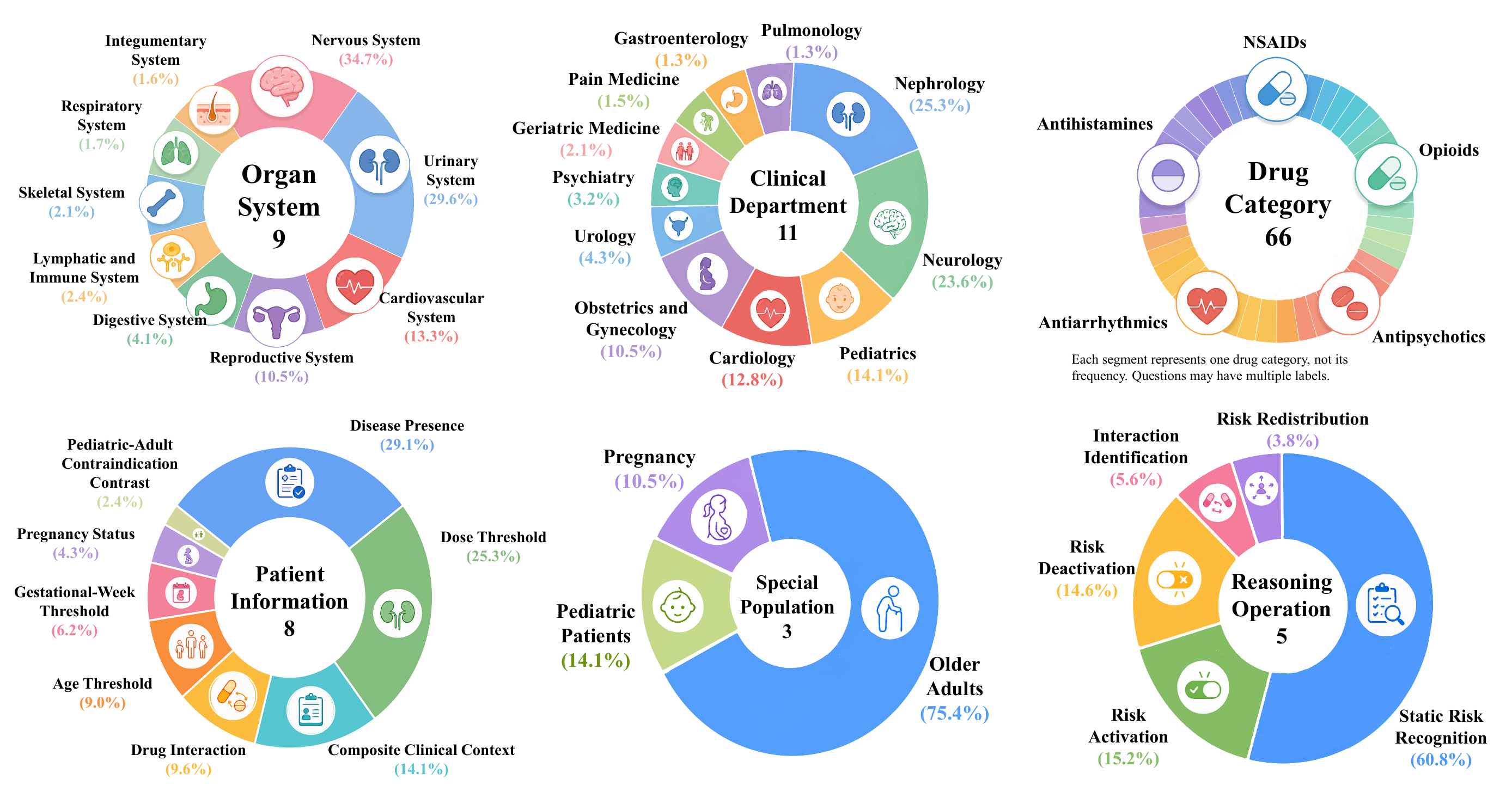}
    \caption{Distribution of MedPIC-Bench questions across six annotation dimensions. The benchmark spans 9 organ systems, 11 clinical departments, 66 drug categories, 8 patient-information types, 3 special populations, and 5 reasoning operations. Percentages indicate the share of questions in each category.}
    \label{fig:classification}
\end{figure*}

\begin{table*}[t]
\centering
\setlength{\tabcolsep}{4pt}
\begin{tabular}{@{}>{\raggedright\arraybackslash}m{2.20in}>{\centering\arraybackslash}m{0.82in}>{\centering\arraybackslash}m{1.00in}>{\centering\arraybackslash}m{1.12in}>{\centering\arraybackslash}m{1.02in}@{}}
\toprule
Benchmark & \shortstack{Rule\\grounding} & \shortstack{Controlled\\PI change} & \shortstack{Answer-changing\\pairs} & \shortstack{Full expert\\validation} \\
\midrule
MedGUIDE~\cite{li2025medguide} & $\checkmark$ & $\times$ & $\times$ & $\triangle$ \\
RxSafeBench~\cite{zhao2025rxsafebench} & $\checkmark$ & $\times$ & $\times$ & $\times$ \\
PsiBench~\cite{proulx2026three} & $\checkmark$ & $\times$ & $\times$ & $\checkmark$ \\
MediEval~\cite{qu2026medieval} & $\times$ & $\times$ & $\times$ & $\triangle$ \\
MEDEQUALQA~\cite{ghosh2025medequalqa} & $\times$ & $\checkmark$ & $\times$ & $\times$ \\
MedEinst~\cite{chen2026medeinst} & $\times$ & $\checkmark$ & $\checkmark$ & $\triangle$ \\
\textbf{MedPIC-Bench} & $\checkmark$ & $\checkmark$ & $\checkmark$ & $\checkmark$ \\
\bottomrule
\end{tabular}
\caption{Comparison of MedPIC-Bench with representative medical benchmarks. A checkmark, cross, and triangle indicate full support, no support, and partial support, respectively. PI denotes patient information. Rule grounding indicates that gold decisions are derived from explicit clinical guidelines or medication-safety rules. Answer-changing pairs are linked cases in which the gold answer changes. Full expert validation indicates item-level review of the complete benchmark; partial validation denotes subset or indirect review.}
\label{tab:benchmark-comparison}
\end{table*}

We transform the collected recommendations into a common rule schema that records the medication or class, target population, relevant patient information, recommended action, and applicable exceptions. Candidate rule fields are first drafted with LLM assistance from source tables, PDFs, and natural-language recommendations. Each draft is then verified field by field against its original source and finalized manually. The finalized records preserve the action specified by each source recommendation, such as avoiding a medication or combination, adjusting a dose, or selecting an alternative treatment. This process provides a source-traceable ground truth for each generated question.

We use deterministic generation scripts to construct questions from the verified rules. The clinical setting and question wording vary across items to reflect the source recommendation and decision type, while the rule-relevant patient information and ground truth remain controlled. Each item comprises a patient vignette, a medication-safety question, and a set of answer options. Distractors are selected from other medications in the source tables or from commonly used medications that do not trigger the target rule in the stated context. A ``None of the above'' option is included when appropriate.

GF questions present a fixed patient case and ask which recommendation applies. They cover medication avoidance, interaction checking, renal dose adjustment, and treatment-selection decisions. CF questions are generated from the same rule representation but change targeted patient information, such as age, pregnancy status, kidney function, disease presence, or an indication-specific exception. Within each linked comparison, the medication candidates and question are held fixed, while non-target clinical details are preserved whenever possible. The correct answer may therefore change from one or more medications to ``None of the above,'' or from one medication set to another.

All generated questions undergo expert validation against the verified rules. Experts follow a shared protocol that assesses source alignment, the correctness of patient information, the completeness of the ground-truth answers, the validity of the distractors, and the consistency of each counterfactual comparison. Questions that fail any criterion are excluded, including ambiguous items, counterfactual comparisons that do not produce a valid contrast, and questions that remain unresolved after expert discussion. Initial inter-reviewer agreement on item validity was 85.3\%. After processing, MedPIC-Bench contains 467 questions, comprising 284 GF and 183 CF questions. To support fine-grained analysis, each question is annotated by organ system, clinical department, drug category, special population, patient-information type, and reasoning operation, as shown in Figure~\ref{fig:classification}. Initial labels are derived from the structured rules and source metadata and then manually reviewed. Table~\ref{tab:benchmark-comparison} further situates MedPIC-Bench among representative medical benchmarks by comparing their source grounding, counterfactual design, and expert validation.

\section{Experiments}

\begin{table*}[t]
\centering
\small
\setlength{\tabcolsep}{4.5pt}
\begin{tabular}{lrrrrrrr}
\toprule
Model & Overall $\uparrow$ & GF $\uparrow$ & CF $\uparrow$ & $\Delta_{\mathrm{GF-CF}}$ & Activation $\uparrow$ & Deactivation $\uparrow$ & Pair $\uparrow$ \\
\midrule
\rowcolor{black!10}\multicolumn{8}{l}{\textit{Medical-specific (M)LLMs}} \\
HealthGPT-Pro-8B & 42.6 & 51.8 & 28.4 & 23.3 & 42.3 & 20.6 & 5.6 \\
MedGemma-4B & 39.6 & 45.4 & 30.6 & 14.8 & 46.5 & 10.3 & 5.6 \\
Lingshu-7B & 43.3 & 50.7 & 31.7 & 19.0 & 39.4 & 32.4 & 6.7 \\
Hulu-Med-7B & 42.0 & 48.2 & 32.2 & 16.0 & 36.6 & 38.2 & 14.6 \\
HealthGPT-Pro-4B & 36.0 & 37.7 & 33.3 & 4.3 & 36.6 & 41.2 & 11.2 \\
HuatuoGPT-o1-8B & 44.3 & 50.0 & 35.5 & 14.5 & 49.3 & 20.6 & 13.5 \\
Fleming-R1-7B & 50.7 & 59.9 & 36.6 & 23.2 & 53.5 & 29.4 & 12.4 \\
Baichuan-M2-32B & 47.8 & 54.6 & 37.2 & 17.4 & 49.3 & 30.9 & 11.2 \\
HuatuoGPT-o1-70B & 55.2 & 63.7 & 42.1 & 21.7 & 63.4 & 23.5 & 14.6 \\
Fleming-R1-32B & 60.6 & 69.0 & 47.5 & 21.5 & 59.2 & 44.1 & 20.2 \\
MedGemma-27B-Text & 61.2 & 67.6 & 51.4 & 16.2 & \textbf{76.1} & 32.4 & 23.6 \\
Lingshu-32B & 59.5 & 63.7 & 53.0 & 10.7 & 60.6 & 54.4 & 25.8 \\
\midrule
\rowcolor{black!10}\multicolumn{8}{l}{\textit{General (M)LLMs}} \\
Gemma-3-12B & 39.6 & 44.0 & 32.8 & 11.2 & 57.7 & 7.4 & 3.4 \\
Gemma-3-27B & 40.3 & 44.7 & 33.3 & 11.4 & 53.5 & 10.3 & 5.6 \\
Llama-3.1-8B & 40.0 & 44.4 & 33.3 & 11.0 & 54.9 & 10.3 & 5.6 \\
Llama-3.1-70B & 55.7 & 61.3 & 47.0 & 14.3 & 63.4 & 32.4 & 16.9 \\
GPT-OSS-20B & 59.1 & 66.5 & 47.5 & 19.0 & 52.1 & 48.5 & 24.7 \\
Qwen3.5-9B & 63.8 & 73.9 & 48.1 & 25.9 & 54.9 & 47.1 & 24.7 \\
Qwen3.5-35B-A3B & 68.5 & 77.5 & 54.6 & 22.8 & 70.4 & 45.6 & 30.3 \\
Qwen3.5-27B & 69.4 & 78.2 & 55.7 & 22.4 & 67.6 & 51.5 & 31.5 \\
GPT-OSS-120B & 70.7 & 79.6 & 56.8 & 22.7 & 62.0 & 55.9 & 30.3 \\
\midrule
\rowcolor{black!10}\multicolumn{8}{l}{\textit{Proprietary (M)LLMs}} \\
Gemini-2.5-Pro & 54.4 & 60.6 & 44.8 & 15.8 & 63.4 & 27.9 & 20.2 \\
Claude-Sonnet-4.6 & 64.2 & 74.6 & 48.1 & 26.6 & 62.0 & 35.3 & 21.3 \\
GPT-5.2 & 68.1 & 78.2 & 52.5 & 25.7 & 63.4 & 42.6 & 29.2 \\
DeepSeek-V4-Pro & 71.7 & 81.3 & 56.8 & 24.5 & 67.6 & 50.0 & 32.6 \\
Qwen3.5-Plus & 72.4 & 81.7 & 57.9 & 23.8 & 66.2 & 55.9 & 32.6 \\
GPT-5 & \underline{75.6} & \underline{83.5} & \underline{63.4} & 20.1 & 70.4 & \underline{61.8} & \underline{38.2} \\
Gemini-3.1-Pro & \textbf{80.7} & \textbf{87.7} & \textbf{69.9} & 17.7 & \underline{74.6} & \textbf{77.9} & \textbf{48.3} \\
\bottomrule
\end{tabular}
\caption{Performance of 28 models on the MedPIC-Bench. Arrows indicate the direction of better performance. $\Delta_{\mathrm{GF-CF}}$ is the GF-minus-CF accuracy gap. Activation and Deactivation are computed over the corresponding reasoning operations, and Pair is the percentage of all linked counterfactual pairs for which both cases are answered correctly. For performance metrics other than the GF--CF gap, best and second-best results are shown in bold and underlined, respectively.}
\label{tab:main-results}
\end{table*}

\subsection{Experimental Setup}

\textbf{Evaluation Models.} We evaluate 28 models spanning medical-specific, general, and proprietary LLMs. The 12 medical-specific models comprise HealthGPT-Pro (4B and 8B) \cite{lin2025healthgpt}, MedGemma (4B and 27B-Text) \cite{sellergren2025medgemma}, Lingshu (7B and 32B) \cite{xu2025lingshu}, Hulu-Med-7B \cite{jiang2025hulu}, HuatuoGPT-o1 (8B and 70B) \cite{zhang2023huatuogpt}, Fleming-R1 (7B and 32B) \cite{liu2026scaling}, and Baichuan-M2-32B \cite{dou2025baichuan}. The 9 general models comprise Gemma-3 (12B and 27B) \cite{gemmateam2025gemma3technicalreport}, Llama-3.1 (8B and 70B) \cite{grattafiori2024llama}, GPT-OSS (20B and 120B) \cite{agarwal2025gpt}, and Qwen3.5 (9B, 27B, and 35B-A3B) \cite{team2026qwen3}. We additionally evaluate 7 proprietary models: Gemini-2.5-Pro \cite{comanici2025gemini}, Gemini-3.1-Pro \footnote{https://deepmind.google/models/gemini/pro/}, Claude-Sonnet-4.6 \footnote{https://www.anthropic.com/news/claude-sonnet-4-6}, GPT-5 \cite{singh2025openai}, GPT-5.2 \cite{singh2025openai}, DeepSeek-V4-Pro \cite{xu2026deepseek}, and Qwen3.5-Plus \cite{team2026qwen3}. The evaluated models include both text-only and multimodal architectures, all of which receive the same inputs.

\textbf{Evaluation Metrics.} We use question-level exact-match accuracy, counting a prediction as correct only when its selected option set exactly matches the ground truth. We report overall, GF, and CF accuracy, together with the GF--CF gap, defined as $\mathrm{Acc}_{\mathrm{GF}}-\mathrm{Acc}_{\mathrm{CF}}$. For CF questions, we additionally report accuracy on risk activation and risk deactivation, as well as pair accuracy over the 89 linked pairs. A pair is correct only when both of its questions are answered correctly. To quantify statistical uncertainty, we report 95\% cluster-bootstrap confidence intervals for the principal aggregate results in the \textit{Uncertainty Estimation} section of the supplementary material. 

\textbf{Experiment Details.} Each model answers every question independently in a zero-shot setting. Members of a CF pair are presented separately without revealing their connection. All models receive the same prompt, which requests a brief rationale followed by the selected option letter or letters in a designated answer field. Open-source inference is conducted on a Linux system with two NVIDIA A800 GPUs, each with 80 GB of memory. We use PyTorch \cite{paszke2019pytorch}, Hugging Face Transformers \cite{wolf2020transformers}, and SGLang \cite{zheng2024sglang} for open-source models, whereas proprietary models are queried through APIs. Further details, including the complete prompt and model-specific inference configurations, are provided in the \textit{Detailed Experimental Setup} section of the supplementary material.

\begin{figure*}[t]
    \centering
    \includegraphics[width=\textwidth]{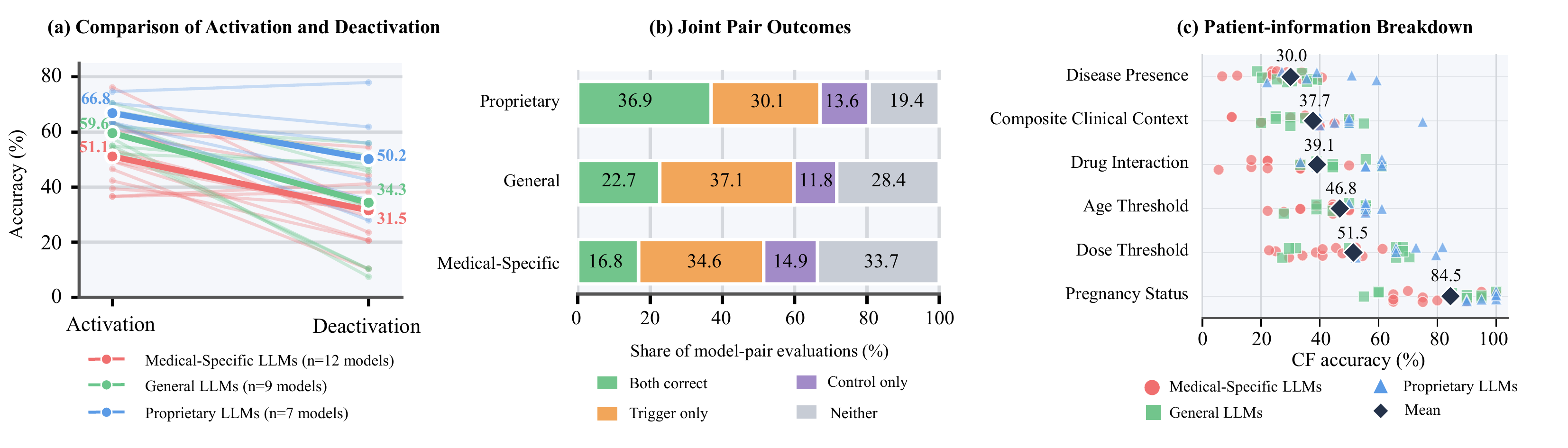}
    \caption{Fine-grained analysis of counterfactual performance. (a) Activation and deactivation accuracy for individual models (thin lines) and model-group means (thick lines). (b) Joint outcomes across 67 matched activation--deactivation pairs, aggregated over all model--pair evaluations within each group. ``Trigger only'' and ``Control only'' indicate that only the rule-triggering or rule-inactive member is answered correctly, respectively. (c) Model-level CF accuracy across six patient-information types, with black diamonds indicating the corresponding means. Gestational-week thresholds occur only in GF questions, while the pediatric--adult contraindication contrast contains only four CF questions and is excluded from the distributional comparison.}
    \label{fig:fine-grained-analysis}
\end{figure*}



\subsection{Main Results}

\textbf{Guideline-following performance consistently exceeds counterfactual performance.} Table~\ref{tab:main-results} reports the complete results for all 28 models. Mean accuracy falls from 63.6\% on GF questions to 45.1\% on CF questions, and every model declines. The 18.5-point gap remains positive under cluster bootstrap (95\% CI [12.6, 24.0] points), showing that the aggregate decline is stable to resampling of question clusters. Together with the model-wise decline, this indicates that the deficit is not confined to a small subset of models or rule clusters. Fixed-case accuracy therefore does not fully reflect model performance when the same knowledge must be applied conditionally as patient information changes. Mean pair accuracy is only 20.0\%. Because pair accuracy requires both linked cases to be answered correctly, it distinguishes consistent conditional application from isolated success on one member of a comparison. A small GF--CF gap does not necessarily indicate robustness because it can also result from low accuracy on both task families. Even Gemini-3.1-Pro, the strongest CF model, solves only 48.3\% of linked pairs. Conversely, HealthGPT-Pro-4B has the smallest GF--CF gap but low CF and pair accuracy. These contrasting profiles show why the GF--CF gap must be interpreted jointly with absolute CF and pair performance. Inspection of the pairs solved by Gemini-3.1-Pro suggests that its advantage comes from explicitly binding each judgment to the rule's applicability conditions, rather than retaining familiar drug warnings.

\textbf{Models activate medication-safety rules more reliably than they withdraw them.} Across the 28 models, mean accuracy falls from 57.7\% on activation to 37.1\% on deactivation, with 25 models exhibiting this asymmetry. Figure~\ref{fig:fine-grained-analysis}(a) shows that the pattern holds across all three model groups rather than being driven by one class of models. Figure~\ref{fig:fine-grained-analysis}(b) further separates the joint outcomes of matched comparisons. Trigger-only outcomes, in which only the rule-active case is answered correctly, are more than twice as frequent as control-only outcomes in every group. Because the paired cases differ through a controlled change in patient information, this imbalance reflects a directional difficulty in withdrawing a warning rather than generic question difficulty alone. Models therefore preserve medication warnings more reliably than they retract them. The response analysis below examines why this asymmetry persists even when changed patient information is recognized.

\textbf{Medical specialization improves performance but does not resolve counterfactual failures.} Medical-specific LLMs average 38.3\% on CF questions, compared with 45.5\% for general LLMs and 56.2\% for proprietary LLMs. Pair accuracy follows the same ordering at 13.8\%, 19.2\%, and 31.8\%, respectively. These group-level comparisons conflate model scale, architecture, and training data, but MedGemma-27B and Gemma-3-27B provide a closer within-family comparison. MedGemma raises CF accuracy from 33.3\% to 51.4\% and pair accuracy from 5.6\% to 23.6\%, indicating that medical adaptation can yield substantial gains. Yet its activation accuracy reaches 76.1\% while deactivation remains at 32.4\%. Medical adaptation therefore improves absolute performance without correcting the central asymmetry. Representative errors further suggest that the remaining deficit is not simply a lack of medical knowledge. Medical-specific models often preserve a medically plausible risk while broadening its threshold, overlooking an indication-specific exception, strengthening the recommended action, or applying a class-level warning inconsistently. Specialization may reinforce familiar drug-risk associations without ensuring control over a rule's threshold, exceptions, action strength, or drug-class scope.

\subsection{Sensitivity across Dimensions}
Figure~\ref{fig:fine-grained-analysis}(c) shows substantial variation across patient-information types. Mean CF accuracy ranges from 30.0\% for condition-triggered rules to 84.5\% for pregnancy status. Composite clinical contexts (37.7\%) and drug interactions (39.1\%) are also difficult, while age thresholds (46.8\%) and renal or dose thresholds (51.5\%) fall between these extremes. Table~\ref{tab:department-cf} shows a consistent gap between departments, with obstetrics and gynecology performing best and neurology and cardiology proving most difficult. Rationale analysis provides a common explanation for both patterns. Models are most reliable when an explicit patient attribute maps directly to a familiar contraindication, as in many pregnancy questions. They struggle when patient information must delimit the scope or strength of a broader safety rule, or deactivate it entirely. The central difficulty therefore lies not in recalling drug risks, but in controlling when and how those risks apply. Results for the remaining dimensions are reported in the \textit{Full Analysis across Dimensions} section of the supplementary material.

\begin{table}[t]
\centering
\small
\setlength{\tabcolsep}{3pt}
\begin{tabular}{@{}lrrr@{}}
\toprule
Department & Medical & General & Proprietary \\
\midrule
Neurology & 26.4 & 31.6 & 40.7 \\
Cardiology & 27.9 & 33.0 & 42.9 \\
Pediatrics & 42.9 & 47.4 & 54.9 \\
Nephrology & 41.1 & 53.0 & 67.2 \\
Obstetrics \& Gynecology & 80.0 & 81.1 & 96.4 \\
\bottomrule
\end{tabular}
\caption{Mean CF accuracy (\%) by model group for clinical departments containing at least 20 CF questions.}
\label{tab:department-cf}
\end{table}

\subsection{Rule-Deactivation Failure Analysis}

\begin{figure}[t]
\centering
\includegraphics[width=\columnwidth]{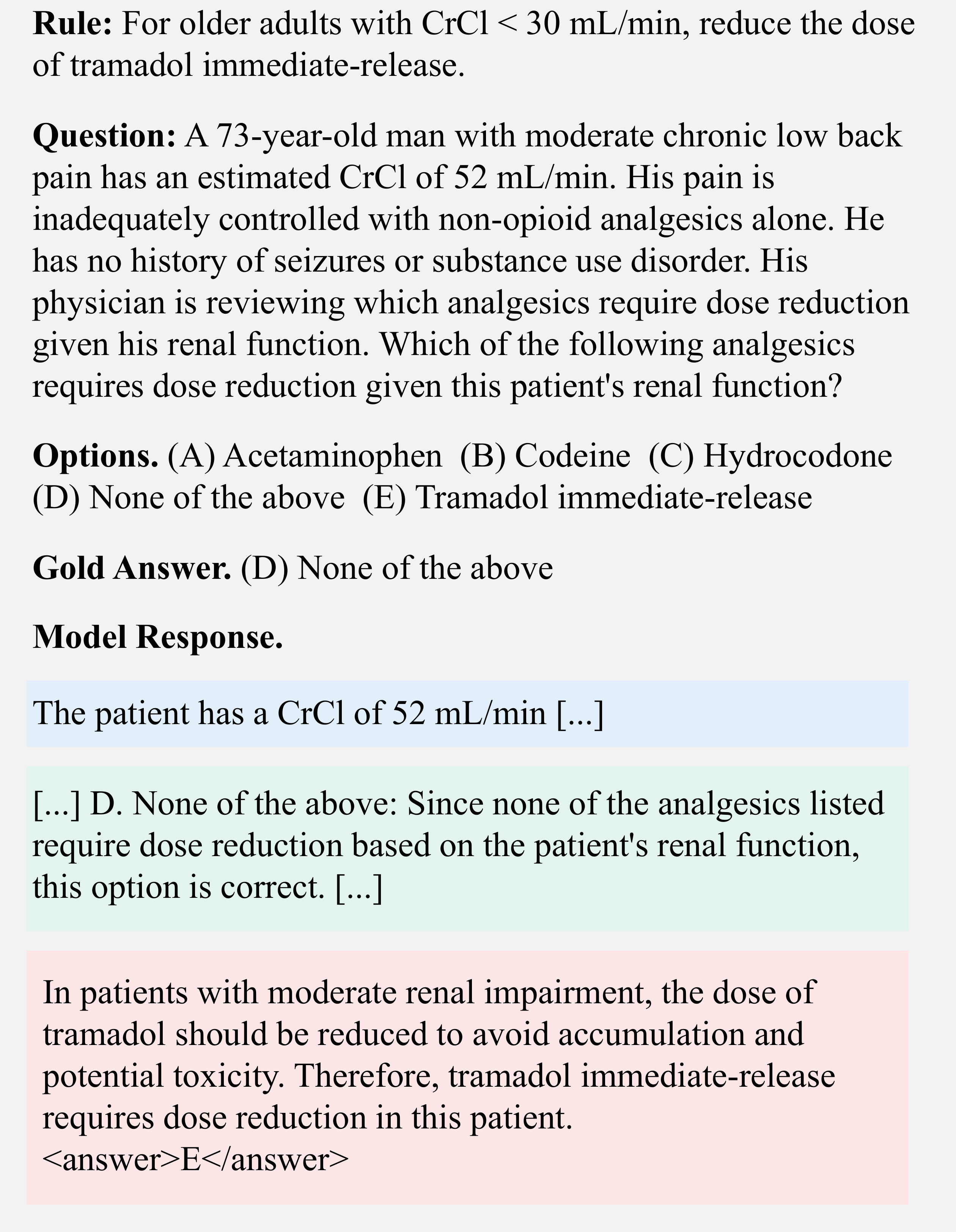}
\caption{A representative rule-deactivation failure by Lingshu-32B. The model initially identifies the rule-consistent answer but then reverts to a broader concern about tramadol accumulation.}
\label{fig:case-study}
\end{figure}

To distinguish failures of context recognition from failures of rule application, we analyze model responses to incorrect deactivation questions. Models mention the changed patient information in 85.6\% of these responses, yet retain the same medication selection in 69.8\%. Figure~\ref{fig:case-study} illustrates a clear instance of this disconnect, in which a correct intermediate judgment is overridden before the final answer.

Lingshu-32B recognizes the patient's CrCl of 52 mL/min and explicitly identifies ``None of the above'' as correct. Yet its final reasoning falls back on a generic concern about tramadol accumulation in renal impairment, extending the verified $<30$ mL/min rule to this patient. The model therefore uses the relevant patient information and reaches a rule-consistent judgment, but allows a broader drug-risk prior to override it before the final answer. This example reflects a broader pattern in which models acknowledge changed patient information without allowing it to govern the final judgment. Reducing such failures may require models to bind each safety judgment to its triggering condition and recheck that condition before producing the final answer.

\section{Conclusion}

Using MedRule2Pair, our source-grounded construction pipeline, we created MedPIC-Bench to test whether LLMs apply medication-safety recommendations within their intended patient scope. Its guideline-following questions measure rule application in fixed cases, while controlled counterfactual pairs test whether judgments change when patient information alters rule applicability. Across 28 LLMs, performance on static questions consistently overstates counterfactual performance. Models activate safety rules more reliably than they withdraw them and frequently preserve medication warnings even after acknowledging that the triggering condition is absent. Medical-specific LLMs offer no consistent robustness advantage, and errors are most pronounced when patient information must constrain, weaken, or deactivate a familiar drug-risk association. These results identify control over rule applicability, rather than risk recall alone, as a central weakness in current medication-safety reasoning. MedPIC-Bench provides a source-traceable testbed for evaluating and developing models whose clinical judgments remain tied to the patient information that justifies them.

\section*{Ethical Statement}

MedPIC-Bench is intended for research evaluation, not clinical decision support. Its synthetic vignettes reduce source recommendations to controlled rule-applicability judgments and do not capture the uncertainty, patient preferences, local guidance, or clinician judgment involved in real prescribing. Results should not be interpreted as treatment recommendations or evidence of readiness for clinical deployment. The benchmark contains no patient records or personally identifiable information, and its release will include source provenance and documentation of these limitations.

\bibliography{aaai2027}


\end{document}